\documentclass[review]{elsarticle}

\usepackage{lineno,hyperref}
\usepackage{subfig}
\usepackage{graphicx}
\usepackage[table]{xcolor}
\usepackage{amsmath,amsthm,amsfonts}
\usepackage{algorithm} 
\usepackage[noend]{algpseudocode} 

\usepackage[justification=centering]{caption}
\usepackage{lineno,hyperref}
 \usepackage{mathptmx}      
\usepackage{latexsym}
\modulolinenumbers[5]

\journal{arxiv}

\begin{document}

\begin{frontmatter}

\title{Local and Global Structure Preservation Based Spectral Clustering}
\author[a]{Kajal Eybpoosh}
\author[b]{Mansoor Rezghi}
\author[a]{Abbas Heydari}

\address[a]{Department of Mathematics, Tarbiat Modares University, Tehran, Iran  }
\address[b]{Department of Computer Science, Tarbiat Modares University, Tehran, Iran }




\begin{abstract}
Spectral Clustering (SC) is widely used for clustering data on a nonlinear manifold.   SC aims to cluster data by considering  the preservation of the local neighborhood structure on the manifold data.   This paper extends Spectral Clustering to Local and Global Structure Preservation Based Spectral Clustering (LGPSC)  that incorporates both global structure and local neighborhood structure simultaneously.  For this extension,  LGPSC proposes two models to extend local structures preservation to local and global structures preservation: Spectral clustering guided Principal component analysis model and Multilevel model.  Finally,  we compare the experimental results of the  state-of-the-art methods with our two models of LGPSC on various data sets such that the experimental results confirm the effectiveness of our LGPSC models to cluster nonlinear data.
\end{abstract}

\begin{keyword}
Spectral clustering\sep Local structure  \sep Global structure  \sep Similarity

\end{keyword}

\end{frontmatter}


\section{Introduction}
\label{intro}

Clustering is an important task in data analysis,  machine learning,  and data mining.  The primary purpose of clustering is to group data points into corresponding categories according to their intrinsic similarities               
  \cite{B2, F-1, L}.   Most of the clustering approaches can be categorized as local or global.  Among existing clustering methods,  techniques based on applying spectral clustering to a similarity matrix have become extremely popular due to their both local and global variants as follows: 
 \begin{enumerate}
\item Local spectral clustering-based approaches such as  classical spectral clustering algorithm uses kernel functions \cite{D0} or k-nearest neighbors (KNN) \cite{Zu},  scalable spectral clustering with cosine similarity \cite{C1},  Local Subspace Affinity (LSA) \cite{Y0},  Spectral Local Best-fit Flats (SLBF) \cite{Zt},  and \cite{Zz} use locality of each data point to build a similarity between pairs of data points. 
\item  Global spectral clustering-based approaches such as Spectral Curvature Clustering (SCC) \cite{C2},  Sparse Subspace Clustering (SSC) \cite{E1,So},  Low-Rank Subspace Clustering (LRSC) \cite{Fa},   Scalable Sparse Subspace Clustering by Orthogonal Matching Pursuit (SSC-OMP) \cite{Y},  and  Sparse Subspace Clustering-matching pursuit (SSC-MP) \cite{Ts}  algorithms that build similarities between data points using global information.
  \end{enumerate}
The idea of spectral clustering \cite{A, Wz,Ss1} is often of interest to cluster a set of $n$ data points on a nonlinear manifold into a finite number of clusters.  This algorithm is widely used in various fields,  such as image segmentation \cite{Z},  speaker diarization \cite{Wa},  and multi-type relational data \cite{Lo}.  Spectral clustering  (SC) is a general graph-theoretic framework.  SC separates the clustering task into two main steps: first converts a data set into a graph or a data similarity matrix such that clustering results of spectral clusters are sensitive to conversion similarity matrices or graphs,  and then applies a graph cutting method to identify clusters.   We know that the similarity matrix has an important role in clustering performance.  So,  most spectral clustering methods are designed to obtain a high-quality graph from the data set \cite{N1,Ze}.  However, there are still some problems to limit the spectral clustering method performance,  which includes some of the proposed spectral clustering-based algorithms only consider the local similarity of data,  some of the methods consider the global similarity of data separately and overlook the local similarity of the data.   To address this,  \cite{We} has merged the local method \cite{H0} and the global method \cite{Zh} to learn similarities of the original data in the kernel space.  Furthermore,  \cite{We} uses low-rank constraint to make the adaptive graph to achieve the purpose of one-step clustering.  Compared to spectral clustering,  the computational cost of the method in \cite{We} is very expensive,  which is not suitable for large data sets.\\

To solve these problems,  we propose a novel spectral clustering method called Local and Global Structure Preservation Based Spectral Clustering (LGPSC).  The LGPSC algorithm simultaneously considers preserving the local  and the global structure of data to provide comprehensive similarities for nonlinear clustering tasks. The remainder of this paper is structured as follows.  Some preliminaries are reviewed in section 2.  The proposed method (including two models) is introduced in section 3.  Section 4 presents experimental  results for several data sets.  The last section is devoted to some conclusions.

\section{Spectral Clustering }
\label{sec:1}
  Spectral clustering  (SC) has became a popular clustering technique due to the pioneering works \cite{M,N,S} at the beginning of the century.   Spectral clustering is a general graph-theoretic framework based on two main steps:
 first converts data points into a data similarity matrix or a graph  and then uses a graph cutting method to identify clusters.  Here we briefly outline the algorithm of spectral clustering. \\

For given data set $X = [x_1, x_2, ..., x_n] \in \mathbb{R}^{m\times n}$   Spectral clustering  (SC) algorithm constructs local similarity matrix $ W=(W_{ij})\in \mathbb{R}^{n\times n} $ that $ W_{ij} $ measures the similarity between two points $  x_{i}$ and $x_{j} $.  The similarities between any two points in $ W $ will be assigned to the value of a Gaussian kernel.  If $ x_{j} $ is a  $ k $ nearest neighbor of $ x_{i} $,  $ W_{ij} $  is computed as follows:
\begin{equation}
W_{ij}=\exp(-d(x_i,x_j)^{2}/\sigma^{2})
\end{equation}

Subject to the constraint $ W_{ij}=0 $ if $ x_{j} $ is not a $ k $ nearest neighbor of $ x_{i} $.  \\

\[ W_{ij} = \begin{cases}  e^{\frac{-dis(x_i,x_j)^{2}}{\sigma ^{2}}} & \quad  x_j \in knn( x_i)\\ 0 & \quad  otherwise\end{cases} \]

By this similarity matrix,  Spectral clustering  (SC) method tries to find embedding vectors $ y_i $ for  each $ x_i $ such that minimize the following object function:

\begin{equation}\label{sp}
\min\;\; tr(Y^T LY)\;\;\; s.t.\;\;\;Y^TY=I
\end{equation}

where $ Y=[y_1, y_2, ..., y_n]^T \in \mathbb{R}^{n\times d} $,   $ d $ is the number of clustering,   $ L=D-W $ is a  Laplacian matrix and $ D=diag( d_1,d_2,\cdots ,d_n ) $,   $ d_{ii}=\sum_{j=1}^{n}W_{ij} $.  It is known that the $ k $ eigenvectors of $ L $ associated with the $ d $ smallest eigenvalues are the solution of the problem.  It is easy to show that $ tr(Y^T LY)=\frac{1}{2} \sum_{i,j=1}^{n} w_{ij}\Vert y_i-y_j\Vert^{2}$. So, the Spectral Clustering (SC) model could be presented as follows:

\begin{equation}
\min\;\;\sum_{i,j=1}^{n} w_{ij}\Vert y_i-y_j\Vert^{2}\;\;\; s.t.\;\;\;Y^TY=I
\end{equation}

\section{Local and Global Structure Preservation Based Spectral Clustering (LGPSC)}
\label{sec:2}

This section proposes a new clustering method called Local and Global Structure Preservation Based Spectral Clustering (LGPSC), a clustering method based on modified Spectral Clustering (SC) to preserve the global and local structure of the nonlinear manifold data. To maintain the global structure of manifold data, the Local and Global Structure Preservation Based Spectral Clustering (LGPSC) technique provides two new models:

\begin{itemize}
\item 
\textbf{First model:} The LGPSC(SC-PCA) technique combines Principal component analysis (PCA) \cite{Du,J} as a classical low-dimensional representation method and Spectral Clustering method.  This model aims to  preserve local and global structure by combining the SC’s similarities in local manifold structure and PCA’s good preservation of the data's global structure simultaneously. 
\item 
\textbf{Second model:} The LGPSC(Multilevel) method uses the arithmetic mean of  each input point along with its k nearest neighbors in the second model.   This model of the LGPSC algorithm employs  the arithmetic means to calculate the similarity between the local neighborhood of each input point and the local neighborhood of other input points,  which makes the global properties to be considered in calculating the similarity matrix.  
 \end{itemize}
  Our proposed methods based spectral clustering can have the following benefits:
\begin{enumerate}

\item Our novel clustering method via exploring the nonlinear similarity relationship of data points and arithmetic mean points to capture the nonlinear and inherent correlation.  
\item Our model  further can preserve the local and global structure of original data to provide comprehensive information for clustering tasks.
\item It also has a compact closed-form solution and can be efficiently computed.
\end{enumerate}

Let $X = [x_1, x_2, ..., x_n] \in \mathbb{R}^{m\times n}$ be a $ m\times n $ data matrix located on the  manifold $ M $  with $ n $ data points and $m $ features,  $x_i$ be the $i$th row of $ X $  which is used to represent the $i$th data point.   
We know that local similarities of data are of great significance for clustering analysis and nonlinear relationship modeling.   Furthermore,  preserving local structure of data is important to spectral clustering.  For this aim,  we first intend to preserve the local structure by using the local objective function can be described as the objective function mentioned in spectral clustering:
\begin{equation}\label{L}
\dfrac{1}{2} \sum_{i,j}\Vert y_i-y_j\Vert^{2}W_{ij}=tr(Y^T LY)
\end{equation}
where $ Y=[ y_1, ... , y_n]^T \in \mathbb{R}^{n\times d} $,  $ L=D-W $,  and $ D_{ii}=\sum_{j}W_{ij} $.\\

In the local method mentioned,  the local neighborhood relationships of these points  $\lbrace x_i\rbrace_{i=1}^{n}$ are considered by using $ k $ nearest neighbors to build similarity matrix $ W $.  Despite that,  it is sensible to consider the global similarity in the objective function.  one way is  selecting the large $ k $ to compute the similarity matrix.  But this viewpoint does not give appropriate results because this way ignores the assumption that the local neighborhood of a data point on the manifold can be well approximated by the k-nearest neighbors of the point.  In the following,  we propose two models that extend the SC method to the LGPSC method.

\subsection{Spectral Clustering Guided Principal Component Analysis (SC-PCA) Model }
 Also, we propose another way that integrates Local spectral clustering (SC) and Principal component analysis (PCA) into a Spectral clustering guided PCA model (SC-PCA) that incorporates both global structure and local neighborhood structure simultaneously while performing clustering.  Combining the objective function of PCA  and the objective function of  SC into a single objective function,  we compute Y and projection U by solving the following 
\begin{equation}
\min_{U,Y}\;\; \sum_{i,j}\Vert x_i-Uy_i\Vert^{2}+\alpha \Vert y_i-y_j\Vert^{2}W_{ij}\;\;\; s.t. YY^T=I
\end{equation} 
where  $U=[u_1,...,u_d]\in \mathbb{R}^{ m\times d}  $ $ \alpha \geq 0$ is a parameter determining the contribution of PCA vs SC. 
SC-PCA has a compact closed-form solution as follows:

\begin{equation}
\min_{Y}\;\; tr(Y^T (-X^TX+\alpha L)Y)
\;\; s.t. \;\;Y^TY=I
\end{equation} 
From computational aspect,  it is desirable for above matrix $ -X^TX+\alpha L $ to be semi-positive definite.   But $ -X^TX+\alpha L $ is not semi-positive definite.  We replace $ -X^TX+\alpha L $ with $(1-\beta )(I-\frac{X^TX}{\lambda})+\beta \frac{L}{\zeta} $ which is semi-positive definite and has the same eigenvectors with $ -X^TX+\alpha L $  where $ \lambda $ is the largest eigenvalue of matrix $  X^TX$ and $ \zeta $ is the largest eigenvalue of $ L $ (see  \cite{J}for more information).

For the purpose of clustering based on preserving the global and local structure,  we use the $ d $ eigenvectors associated with the $d $ smallest eigenvalues of the matrix $(1-\beta )(I-\frac{X^TX}{\lambda})+\beta \frac{L}{\zeta} $.  Then we can cluster the data into $ d $ groups by applying k-mean to the $ d $ ordered eigenvectors.

\subsection{Multilevel Model}
In this model,  we propose a novel approach called the multilevel model of the LGPSC technique for entering appropriate global geometric similarities in the objective function of SC.   The approach suggests looking at the aspect of the global clustering problem as a process going through different levels evolving from a fine grain to coarse grain strategy.   The clustering problem enters mean points level by level to a global problem.  The clustering of the global problem is mapped back level-by-level to obtain a better clustering of the original problem by using the multilevel mean points,  i.e. ,  in the first level of multilevel clustering as explained in following,  our algorithm uses initial mean points of data points,  and in the next level,  it uses the mean of initial mean points.   In the following,  we aim to explain the multilevel model of our algorithm in the first level.

The Spectral Clustering method only calculates the similarity of each point $ x_i $ with its nearest points $ x_j \in knn(x_i)$,  i.e., the similarity in each locality is calculated,  but our method uses the arithmetic mean of each point and its nearest points $ z_i=mean(x_i, knn(x_i))$ such that each mean  $ z_i $ represents a locality.   By calculating the similarity between each mean point $ z_i $ and its neighbors $ knn(z_i)$,  the similarities between a locality and its neighbors are calculated.  So we can add these global similarities to the local method by using mean points.

Firstly in this model,  for each point, we form an arithmetic mean of it and its neighboring points so that if the data set has $ n $ points,  then $ n $  arithmetic means are formed up.  That is,   we have $n$ arithmetic mean points  in the form of $z_1,...,z_n$,  where $z_i$ is the mean of the point $x_i$ and its neighbors.  These $n$ arithmetic mean points are considered as a set of $n$  points. 
\begin{equation}
Z=[ z_1, ... , z_n]
\end{equation}

\[z_i=\frac{1}{k+1}\sum \limits_{s \in \lbrace i,knn(i) \rbrace }x_s, \quad knn(i):=\lbrace j \vert x_j \in knn(x_i)  \rbrace\]
 For each point we consider the $ k^{\prime} $  nearest neighbors and  construct a matrix of similarities $ W^{\prime}\in \mathbb{R}^{n\times n} $ that $ W^{\prime} _{ij} $ measures the similarity between two points $  z_{i}$ and $z_{j} $.  The similarities between any two points in $ W^{\prime}  $ will be assigned to the value of a Gaussian kernel.  If $ z_{j} $ is a  $ k^{\prime}  $ nearest neighbor of $ z_{i} $,  $ W^{\prime} _{ij} $  is computed as  $ W$.  

\[ W^{\prime} _{ij} = \begin{cases}  e^{\frac{-dis(z_i,z_j)^{2}}{\sigma ^{2}}} & \quad  z_j \in knn( z_i)\\ 0 & \quad  otherwise\end{cases} \]

We use the following objective function to global cluster these points:

\begin{equation}\label{G}
\frac{1}{2}\sum_{i,j}\Vert t_i-t_j\Vert^{2}W^{\prime}_{ij}=tr(T^T L^{\prime}T)
\end{equation}

We define
\begin{equation}
 T=[ t_1,..., t_n]^T \in \mathbb{R}^{n\times d},  t_i^{T}=\frac{1}{k+1}\sum \limits_{s \in \lbrace i,knn(i) \rbrace }y_s^{T}=h_i^{T}Y
\end{equation}
where $ L^{\prime}=D^{\prime}-W^{\prime} $,   $ D^{\prime}_{ii}=\sum_{j}W^{\prime}_{ij} $ and the elements of $ h_i $ are zero except element that their corresponding $ y_i $ belongs to $ knn(i) $,  in details

\[ (h_i)_{j} = \begin{cases}  1 & \quad  j \in \lbrace knn( i),1\rbrace \\  0 & \quad  otherwise\end{cases} \]

To Combine local formulation of Eq.(\ref{L}) and global formulation of Eq.(\ref{G})
into a single objective function,  we compute $T$ according to $Y$ by solving the following:
\begin{equation}\label{T}
T=\begin{bmatrix} t_1^{T}\\ \vdots \\ t_n^{T}\\ \end{bmatrix}=\frac{1}{k+1}\begin{bmatrix}  h_1^{T}Y\\ \vdots \\ h_n^{T}Y_n\\ \end{bmatrix}=\frac{1}{k+1}HY,\;\;\;H=\begin{bmatrix} h_1^{T}\\ \vdots \\h_n^{T}\\ \end{bmatrix} 
\end{equation}
where

\[ H_{ij} = \begin{cases}  1 & \quad  j \in \lbrace i,knn( i)\rbrace \\0 & \quad  otherwise\end{cases} \]

Therefore,  by substituting Eq.(\ref{T}) instead of $ T $ in Eq.(\ref{G}),  the objective function Eq.(\ref{G}) can be written as follows:

\begin{equation}\label{G1}
\min tr(T^TL^{\prime}T)=\min tr(Y^T H^T L^{\prime} HY)=\min tr(Y^T L^{\prime\prime}Y)
\end{equation}

Since $L^{'}$ is an SPD matrix, So its clear that the matrix $L^{''}$ is also becomes SPD matrix.

We combine local formulation of Eq.(\ref{L}) and global formulation of Eq.(\ref{G1}) into a single objective function.   Therefore,  a nonlinear spectral clustering method based on preserving the global and local structure  is proposed as follows
\begin{equation}\label{LG}
\min tr(Y^T(L^{\prime\prime}+L)Y),
\end{equation}
where  $L^{\prime\prime}+L$ is an symmetric positive matrix.

For the purpose of clustering based on preserving the global and local structure,  we use the $ d $ eigenvectors associated with the $d $ smallest eigenvalues of the matrix $ (L^{\prime\prime}+L) $,  eigenvalues and eigenvectors are computed by solving

\begin{equation}\label{*}
(L^{\prime\prime}+L)u=\lambda u
\end{equation}

There is a mapping $ f:M\longrightarrow \mathbb{R}^{d} $ that we can cluster the data into $ d $ groups by applying k-mean to the $ d $ ordered eigenvectors of the Laplacian matrix.  Let  $ u_1,...,u_{d} $ be the $ d $ eigenvectors of $ L $ associated with its $ d $ smallest eigenvalues,  by applying k-means to the rows of $ [u_1,...,u_{d}] $,  we cluster the manifold $M$ into $ d $ different groups.  
\subsection{Complexity Analysis}

Regarding the computational complexity of our algorithm,  the searching $k$ nearest neighbors takes $ O(mn^2) $ \cite{Ka},  where $ m $ is the dimensionality of input data,  and $ n $ is the number of input data points.  The time complexity of computing the $n$ mean points is $O(kmn)$.   The step of constructing each similarity matrix (or the Laplacian matrix or matrix $(1-\beta )(I-\frac{X^TX}{\lambda})+\beta \frac{L}{\zeta} $)  via knn of all data points takes $ O(mn^2) $ \cite{Ge}.  The computation of $ d $  matrix eigenvectors has time complexity $ O(dn^2) $.  The time complexity of $k$-means is $O(nd^2)\times l  $,  where  $ l $ is the number of k-means iterations.  The overall computational complexity of our algorithm is $  O(nm^2) $. 


\section{Experiments}
In our experiment,  we test our proposed methods by comparing them with six comparison methods on six benchmark data sets.  Moreover,  we evaluate the clustering performance of all methods through 
 Normalized Mutual Information (NMI) \cite{LC},  and Adjusted Rand index (ARI)\cite{Si}  as measurement methods of the clustering qualities.

\subsection{Data sets}
To effectively evaluate all methods,  we chose six data sets with the different sizes and classes:

\begin{enumerate}
\item ORL Face Data Set\footnote{https://scikit-learn.org/stable/datasets/index.html} is composed of $10$ different images of each $40$ distinct persons.  The size of each image is $92\times 112$,  with $256$ grey levels per pixel. In our experiment,  each face image was resized into $64 \times 64$. \\

\item Handwritten Digits  Data Set\footnote{https://scikit-learn.org/stable/datasets/index.html} consists of 8-bit gray-scale images of digits from $”0”$  to $”9”$.  There are about 180 examples for each class.  Each image is centered on an $8\times 8$.\\

\item IRIS Flower Data Set\footnote{https://scikit-learn.org/stable/datasets/index.html} contains $3$ objects.  It consists of $50$ samples with four features from each of three objects.\\

\item MNIST Data Set\footnote{https://scikit-learn.org/0.19/datasets/mldata.html} consists of 8-bit gray images of digits from ”0” to ”9”.  The size of each digit image is 28×28.  \\

\item Wine Data Set\footnote{https://scikit-learn.org/stable/datasets/index.html} consists of a feature matrix with 178 samples as rows and 13 feature columns,  and a target matrix with a 1-dimensional array of the 3 class labels 0, 1, 2.\\

\item COIL20 Data Set\footnote{http://www.cad.zju.edu.cn/home/dengcai/Data/MLData.html} is a database of gray images of 20 objects.  It contains 1440 toy images.  The size of each image is 32×32 pixels. \\
\end{enumerate}

Furthermore,  more details are listed detailedly in Table \ref{tab1}.

\begin{table}[h!]
\centering 
\caption{ Data descriptions: Number of Data points ($ m $),
Number of Dimensions ($n$), Number of Clusters ($ L $).}
{{ \begin{tabular}{c c c c }
  \hline\hline 
Data\; set&$ n $&$ m $&$ d$\\ 
  \hline
ORL &400&4097&40\\  
  \hline  
IRIS &150&4&3\\
  \hline
Handwritten&1797&64&10\\
  \hline
WINE&178&14&3\\ 
  \hline
MNIST&1000&784&10\\
  \hline
COIL20&1440&1024&20\\  
  \hline
\end{tabular}}\label{tab1}}
 \end{table} 
 
\subsection{Comparison methods}
To evaluate the performance of clustering how can be improved by our proposed approaches,  we compared the results of the following algorithms:
\begin{itemize}
\item $k$-Means Clustering \cite{Har} Algorithm:  $k$-means is a clustering method which first randomly selects k initialization as clustering centers and assigns remainder samples to corresponding clusters,  and then iteratively updates until convergence.
\item Spectral Clustering (SC)  \cite{M,N,S} Algorithm: SC is a graph-based clustering that converts clustering tasks into graph division problems.  A data set can be presented as a nonlinear graph or similarity matrix such that  each point of data set can be regarded as a vertex of a graph,  and the edge lengths of a graph can be measured by similarities among points.  Then it uses a graph cutting method to identify clusters.
\item  Sparse Subspace Clustering (SSC) \cite{E1} Algorithm: The SSC algorithm applies spectral clustering to a similarity matrix obtained by finding a sparse representation of each point in terms of other data points.  SSC uses  the $l_1$-norm minimization problem to find the sparse representation.  When  the number of data points is large,  or the data is high-dimensional, the computational complexity of SSC quickly becomes prohibitive.  
\item Scalable Sparse Subspace Clustering by Orthogonal Matching Pursuit (SSC-OMP) \cite{Y} Algorithm: The SSC-OMP algorithm replaces the $l_1$-norm minimization  by the orthogonal matching pursuit (OMP) algorithm \cite{Pa} to find a sparse representation in the SSC method. 
\item Sparse Subspace Clustering-matching pursuit (SSC-MP) \cite{Ts}  Algorithm:  SSC-MP employs the matching pursuit matching (MP)  \cite{Fr,Ma} algorithm  instead of the OMP algorithm  to compute sparse representations of the data points. Although the existing SSC methods already have good theoretical and practical contributions,  their  computational cost is very expensive.
\item Scalable Spectral Clustering with Cosine Similarity \cite{C1} Algorithm: The scalable spectral clustering with cosine similarity algorithm provides a scalable implementation of the spectral clustering algorithm in the special setting of cosine similarity by exploiting the product form of the similarity matrix. Let $\mathbb{R}^{m \times n}$ be a data set of $m$ points in $\mathbb{R}^n$. The scalable spectral clustering with cosine similarity algorithm uses the left singular vectors of $D^{-1 / 2} X$ (where $D=\operatorname{diag}\left(X\left(X^T 1\right) 1\right)$ ) instead of Laplacian matrix eigenvectors.
\end{itemize}
\subsection{Experimental result analysis}
\begin{center}
 \begin{figure}
 \includegraphics[width=14cm]{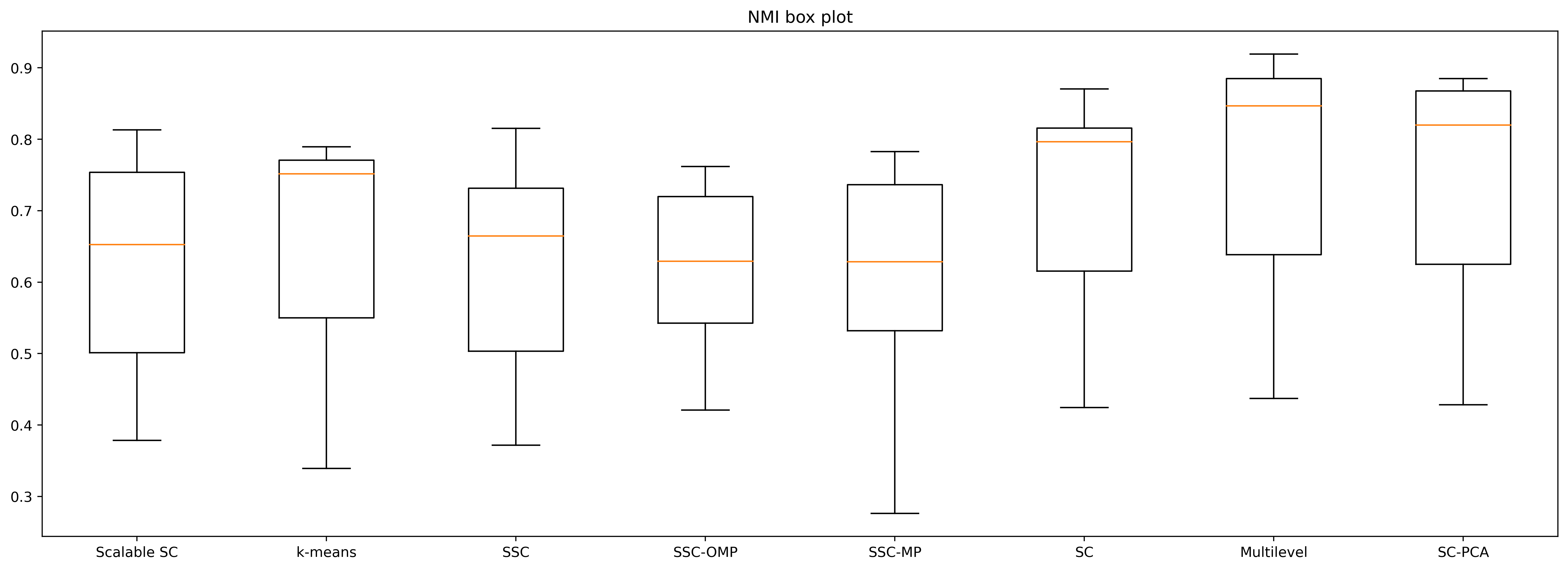}\centering\caption{NMI Box Plot.}\label{c1}
\end{figure}
\end{center}

\begin{center}
 \begin{figure}
 \includegraphics[width=14cm]{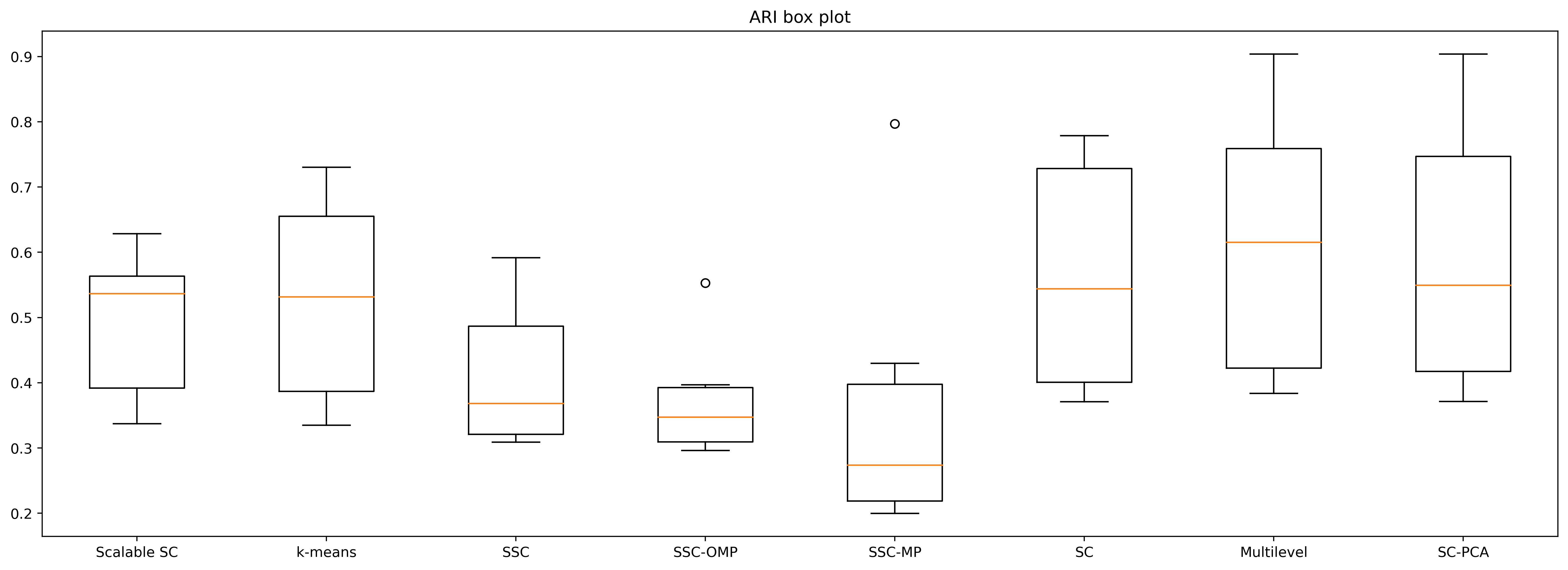}\centering\caption{ARI Box Plot.}\label{c2}
\end{figure}
\end{center}

All clustering results of six data sets are shown in Table \ref{tab12} and Table \ref{tab13},  where the maximum in each column is denoted as bold.  From these tables,  we have the following observations.   We can observe that our proposed methods achieve the best clustering performance.  More specifically,  Table \ref{tab123} shows that the results of our proposed methods increase by $4$ percent at least,   compared with all above mentioned six methods,  in terms of  the clustering metrics including NMI (Figure \ref{c1}) and ARI (Figure \ref{c2}) on six data sets.  The reason is perhaps that our proposed methods make full use of the comprehensive information on both global and local structure information.  \\

\begin{table}[h!]
\centering 
\caption{NMI values for clustering methods across six data sets.}
{ \begin{tabular}{c c c c c c c c c}
  \hline\hline 
  & &&&&&& \multicolumn{2}{c}{LGPSC} \\
   
 Data\; set&Scalable SC&$k$-means&SSC&SSC-OMP&SSC-MP&SC&Multilevel&SC-PCA\\ 
\hline 
   ORL&0.8131&0.7895&0.8154&0.7620& 0.7510&0.8072&\bf{0.8267}&0.8110\\
     \hline
   IRIS &0.5898&0.7586&0.6572&0.6843&0.7827&0.7859&\bf{0.9192} &0.8850\\
      \hline      
  Handwritten&0.7152&0.7449&0.6718&0.5322&0.5643&0.8705&\bf{0.8912} &0.8808\\
    \hline         
    WINE&0.3785&0.3391&0.3719&0.4210&0.2762&0.4244&\bf{0.4371}&0.4283\\
     \hline     
  MNIST&0.4718&0.4848&0.4521&0.5740& 0.5213&0.5587&\bf{0. 5755}&0.5631\\
  \hline
    COIL20&0.7665&0.7749&0.7513&0.7317&0.6926&0.8183&\bf{0.8665}&0.8285\\
      \hline
        \end{tabular}}\label{tab12}
 \end{table}

\begin{table}[h!]
\centering 
\caption{ARI values for clustering methods across six data sets.}
{ \begin{tabular}{c c c c c c c c c}
  \hline\hline 
  & &&&&&& \multicolumn{2}{c}{LGPSC} \\
 Data\; set&Scalable SC&$k$-means&SSC&SSC-OMP&SSC-MP&SC&Multilevel&SC-PCA\\ 
\hline 
   ORL&0.5348&0.4334&0.5210&0.2960&0.2455&0.4525&\bf{0.5374}&0.4540\\
     \hline
   IRIS &0.5388&0.7302&0.5917&0.5528&0.79687&0.7591&\bf{0.9037}&\bf{0.9037}\\
   \hline      
  Handwritten&0.6285&0.6636&0.3837&0.3077&0.3021&0.7787&\bf{0.7809}& 0.7808\\
    \hline         
    WINE&0.3371&0.3711&0.3090&0.3969&0.1997&0.3711&\bf{0.3841}&0.3714\\
     \hline     
  MNIST&0.3441&0.3348&0.3101&0.3149&0.2098&0.3833&0.3836&\bf{0.4053}\\
  \hline
    COIL20&0.5718&0.6301&0.3529&0.3799&0.4296&0.6357&\bf{0.6932}&0.6447\\
      \hline
        \end{tabular}}\label{tab13}
 \end{table}
 We display in Table \ref{tab1234} the running time of the various methods on the six versions of data.  Table \ref{tab123} shows that our method considerably improved the running time of the spectral clustering in most of the data.  On the other hand,  even though our approach lasts more time than some methods to run,  it still gives more accurate results in a reasonable time. 

\begin{table}[h] 
 \centering \caption{Clustering speeds (seconds) of different algorithms.}
 \begin{tabular}{c c c c c c c c c}
  \hline\hline 
  & &&&&&& \multicolumn{2}{c}{LGPSC} \\
   Data\; set&Scalable SC&$k$-means&SSC&SSC-OMP&SSC-MP&SC&Multilevel&SC-PCA\\ 
    \hline
        ORL&4.8252&2.6171&4.4942&63.4916&15.9119&5.3910&8.2249&1.2742\\
      \hline
    IRIS &0.4620&0.0346&0.4350&1.2850&0.3870&0.9770&0.3473&0.2020\\     
      \hline
      Handwritten&4.0612&0.1803&5.4793&13.9928&9.1125&109.9910&44.6239&28.2888\\
        \hline
   WINE&0.1520&0.0360&0.4590&0.4760&0.3500&0.9470&0.5478&0.2517\\  
     \hline
  MNIST&0.4605&0.3005&8.4284&42.7634&18.7320&31.4455&17.2987&7.8027\\
      \hline
    COIL20&7.1413&0.7313&8.8005&110.9863&53.0100&77.8178&37.1621&21.5778\\    
      \hline
        \end{tabular}\label{tab1234} 
    \end{table}
    
\begin{table}[h!]
\centering 
\caption{The average clustering results of six benchmark data sets.}
{ \begin{tabular}{c c c c c c c c c}
  \hline\hline 
& &&&&&& \multicolumn{2}{c}{LGPSC} \\
 Data\; set&Scalable SC&$k$-means&SSC&SSC-OMP&SSC-MP&SC&Multilevel&SC-PCA\\ 
\hline 
   NMI&0.6224&0.6486&0.6199&0.6175&0.5980&0.7108&0.7527& 0.73278\\
     \hline
   ARI &0.4925&0.5272&0.4114& 0.3747&0.3639&0.5634&0.6138&0.5933\\
      \hline    
      Time&2.8503& 0.6499&4.6827&38.8325&14.5839&37.7615&18.0341& 9.8995\\

        \end{tabular}}\label{tab123}
 \end{table}

\section{Conclusion}
In this work,  we have proposed  two novel nonlinear spectral clustering models.   The first model of proposed methods simultaneously uses global  and local structure preservation for providing comprehensive structure information of the data to cluster nonlinear data by employing PCA,  which makes the global properties to be considered in spectral clustering.  Moreover,  we introduce a multilevel model of our proposed technique for the nonlinear clustering that is solved by  arithmetic mean points level by level to obtain a better clustering.  Experimental results on six data sets have shown employing comprehensive information is important to improving clustering performance.  In the future work,  we intend to extend our proposed method into data manifolds with unknown structures by conformal mapping  to  provide a  feature space with known structures.


%
%



\bibliography{mybibfile}

\end{document}